# Probabilistic Semantics and Defaults

Eric Neufeld and David Poole
Department of Computer Science
University of Waterloo
Waterloo, CANADA, N2L 3G1

## Abstract

There is much interest in providing probabilistic semantics for defaults but most approaches seem to suffer from one of two problems: either they require numbers, a problem defaults were intended to avoid, or they generate peculiar side effects.

Rather than provide semantics to defaults, we address the original problem that defaults were intended to solve: that of reasoning under uncertainty where numeric probability distributions are not available. We describe a non-numeric formalism called an inference graph based on standard probability theory, conditional independence and sentences of *confirmation*, where $a$ confirms $b \equiv \mathit{conf}(a,b) \equiv p(a|b) > p(a)$.

The formalism seems to handle the examples from the nonmonotonic literature. Most importantly, the sentences of our system can be verified by performing an appropriate experiment in the semantic domain.

## 1 Introduction

Though default reasoning involves reasoning under conditions of uncertainty, some argue it is not probabilistic reasoning. Reiter and Crisculo [21] distinguish the two by suggesting different interpretations for the word "most". Probabilistic reasoning gives "most" a statistical connotation, whereas default logic gives it a prototypical sense. On the other hand, Poole [18] claims defaults should not be thought of as having any meaning in the sense of "most" or "typical"; they are statements the user is prepared to accept as part of an explanation as to why something may be true.

What, then, does a default mean? Within the default logic camp, we know of no work which provides a semantics for defaults, in the sense that an experiment is described that can be performed in the semantic domain to verify the truth of a default. It is therefore compelling to view defaults as qualitative probabilistic statements where numeric distributions are unavailable. We survey some of these views but note most require numbers, something default reasoning intended to avoid, or have side effects.

Rather than "add semantics" to defaults, we construct a sound non-numeric probabilistic formalism called an *inference graph*. We explore its mathematical properties, then apply it to the standard examples. We conclude with a brief description of the implementation.

## 2 What's in a default?

Poole et al [20] attempt to put both default reasoning and diagnosis under a single umbrella by constructing a system containing a set of facts $F$ known to be true, a set $\Delta$ of defaults, and $g$, a set of (possible) observations which are goals to be proved. Here we assume $F$, $\Delta$ and $g$ are propositional.

If $D$ is a subset of $\Delta$ such that

$$F \cup D \models g, \text{ and } F \cup D \text{ is consistent.}$$

then $D$ is an *explanation* of $g$. This system is based on a theorem prover and can be used in two ways. If $g$ consists of observations known to be true, we interpret $g$ as querying *why g?*, and $D$ is a *diagnosis* of $g$. If $g$ is not known to be true, then $g$ is interpreted as querying *whether g?*, and $g$ is a *prediction* of $F \cup D$. The problem default logic runs into is that there is typically another $D'$ such that $F \cup D'$ predicts $\neg g$; this is known as the *multiple extension problem* and is discussed below. (This is a very abbreviated presentation of default reasoning; for details on implementation and application see [15,14].)

As pointed out in the introduction, defaults appear to have no semantics, and many researchers study the relationship between default reasoning and uncertainty. Rich [22] advocates adding certainty factors to possible hypotheses to fine-tune a default reasoning system and concludes "default reasoning is likelihood reasoning and treating it that way simplifies it". While some argue with her treatment, her conclusions seem to be widely held. Ginsberg [4] pursues this approach.

At the 1987 Workshop on Uncertainty in AI, Grosof suggested defaults are interval-valued probabilities on the entire unit interval. Default inference thus becomes closely related to Kyburg's theory of interval-valued probabilities [8,7].

In [9], McCarthy states non-monotonic sentences can represent statements of infinitesimal probability, but does not go into detail. Pearl explores this in [12]. This interpretation has some problems. Let $e = emu$, $b = bird$, $f = fly$. If

$$p(f|b) \approx 1, p(f|e) \approx 0, p(b|e) \approx 1$$

(i.e., there are some non-bird emus) then from

$$p(f|e) = p(f|be)p(b|e) + p(f|\neg be)p(\neg b|e)$$

Pearl shows $p(f|be) \approx 0$. Then, from

$$p(f|b) = p(f|eb)P(e|b) + p(f|\neg eb)p(\neg e|b)$$

it follows $p(e|b) \approx 0$. But since a prior is always bounded by its conditionals on any evidence and the negation of that evidence, we can show $p(e) \approx 0$.

Default logic also has this "property": from no knowledge at all, we can prove $\neg emu$ by cases from *fly* and $\neg fly$ using the contrapositive forms of the defaults. This introduces the following variant of the "lottery paradox"[8].

**Example 2.1** Suppose kangaroos ($k$) are exceptional because they have a marsupial birth and platypusses ($p$) are exceptional because they lay eggs but dingos ($d$) have no such exceptional traits. If

$$\textit{ozzie-animal} \Rightarrow e \vee k \vee p \vee d,$$

then $p(d|\textit{ozzie-animal})$ is close to one since the disjunction of the other three is close to zero. Default reasoning and circumscription [16] suffer the same problem. Poole [17,3] solves this by explicitly pruning the proof tree with a set of sentences called constraints. However, to do this, you need to know the right answers in advance.

Besides making subclasses vanish, Pearl's $\epsilon$-semantics suffers another problem: in general, it is impossible to go out into a real problem domain and find a set of conditional probabilities infinitely close to one.

Bacchus[1] addresses this issue of practicality and argues for thresholding, that is, that a possible hypothesis stands for a probability greater than some threshold $k > 1/2$. His system allows only a single defeasible inference, since $p(b|a) > k$ and $p(c|b) > k$ do not in general constrain the value of $p(c|a)$ to be greater than $k$.

There seems to be no end to different probabilistic semantics that might be added to defaults or inference rules that might be invented to come up with the right answers for the various examples. We claim it is necessary to ask again what were the original goals of formalisms such as default reasoning, inheritance hierarchies and semantic nets. We should reconsider the original objections to standard probability, and ask whether we solve the problem in a principled way, without the invention of new formalisms.

## 3 Inference graphs

An inference graph is a strictly probabilistic formalism based on standard probability theory, conditional independence and sentences of confirmation. Rather than give rules for *accepting* uncertain conclusions, the inference graph allows us to make inferences about *shifts* in belief.

### 3.1 Confirmation

An interesting mathematical property of logical implication is that knowledge of the consequent increases belief in the antecedent. That is, $a \Rightarrow b$ implies that $p(a|b) \geq p(a)$. Rosenkrantz [23] calls this property *confirmation*, and we will see it has many of the same useful computational properties as other probabilistic formalisms.

Confirmation describes a *shift* in belief; it seems to be the weakest probabilistic property a default *ought* to have. This provides an interesting venue to explore: rather than use knowledge of the form "birds are more likely to fly than not", we consider knowledge of the form "an individual is more likely to fly once we learn that it is a bird".

Consider Nutter's example[10], where in springtime it is not true that most birds fly, since most birds are flightless nestlings. Yet, the information that an individual is a bird inclines us to shift belief in favour of flying.

This also admits an interesting kind of sentence. If we say "Irish Canadians have red hair", we do not mean more than half or almost all Irish Canadians have red hair, even though the stereotype is widely held.

### 3.2 Syntax

An *inference graph* contains four kinds of links, $\Rightarrow$, $\rightarrow$, $\not\Rightarrow$ and $\not\rightarrow$. Links with double arrows are called *logical links* and the others are *probabilistic links*. Each node is labelled with a name or set of names in lower case, for example, *quaker* or *pacifist*. Links are attached to a name or its negation at either endpoint.

### 3.3 Semantics

Nodes in an inference graph denote events. Generally events have two mutually exclusive outcomes, for example *fly* or $\neg fly$. Occasionally an event may have several mutually exclusive outcomes, (not all of which need be specified), for example $\{hawk, dove\}$.

Sentences about confirmation[1] are represented by the four kinds of links in an inference graph:

$$a \rightarrow b \text{ means } p(b|a) > p(b).$$
$$a \Rightarrow b \text{ means } 1 = p(b|a) > p(b).$$
$$a \not\rightarrow b \text{ means } p(\neg b|a) > p(\neg b).$$
$$a \not\Rightarrow b \text{ means } 1 = p(\neg b|a) > p(\neg b).$$

(Note that we insist on strict inequality. This means that links such as $2 + 2 = 3 \Rightarrow$ *sky-is-blue* cannot appear on an inference graph. For the same reason, we also insist all events are possible.)

The topology of the inference graph carries information about independence of events.

**Definition 3.1** If $p(a|b) = (a)$, $a$ and $b$ are *unconditionally independent*.

**Definition 3.2** If $p(a|bc) = p(a|b)$, $a$ is *conditionally independent* of $c$, given $b$.

---

[1] Possibly confirmation is too strong a term where logical implication is not involved. We use confirmation here in the sense of partial confirmation, or relevance.

277

If $a$ is a node, and $b_1, \ldots, b_n$ are the nodes directed into $a$, then $a$ is conditionally independent of all the predecessors of the $b_i$ given the outcomes of the $b_i$.

Thus, an inference graph may be seen as a non-numeric *influence diagram*[25]. We next explore the kinds of inferences about confirmation that we can make.

## 4 The confirmation relation

**Definition 4.1** If $p(a|b) > p(a)$, we also write *conf(a,b)*.

### 4.1 Symmetry

**Lemma 4.2** If $conf(a, b)$, then $conf(b, a)$.

**Proof:** Follows immediately from Bayes' Rule. □

This allows our system to be reversible; if we observe *sneeze* we can confirm *has-cold*. Alternately, if we know that someone has a cold we can predict they will sneeze. Thus we can use the same formalism for prediction and diagnosis.

### 4.2 Negation

**Lemma 4.3** If $conf(a, b)$, then $conf(\neg a, \neg b)$.

**Proof:** By definition, $p(a|b) > p(a)$. Negating both sides yields $p(\neg a|b) < p(\neg a)$. Then $p(b|\neg a) < p(b)$ by Lemma 4.2 and negating again yields $p(\neg b|\neg a) > p(\neg b)$. Another application of Lemma 4.2 yields the desired result. □

Thus, not only does *bird* increase belief in *fly*, *¬bird* increases belief in *¬fly*. An interesting intermediate result is that the "contrapositive" form of a link yields a valid inference, so long as it is made from a single link. This means use of the contrapositive form of a link is valid, but the context of such an inference must be carefully restricted. Inference graphs also explain why default reasoners based on a theorem prover sometime run into difficulties when they apply the contrapositive: they violate independence assumptions.

### 4.3 Logical Inferences

**Lemma 4.4** If $conf(a, c)$ and $conf(b, d)$ where $c$ and $d$ are outcomes of the same random variable, and $a \models b$, then $conf(c, ab)$.

**Proof:** $p(c|ab) = p(c|a) > p(c)$, since sentences of probability hold for logically equivalent propositions. □

Default reasoners produce separate arguments for $c$ and $d$ and attempt to choose among the arguments by appeal to "specificity". Poole [19] calls it preferring the most specific theory and Kirby [6] calls it choosing the most specific extension.

While the default logic view seems to be to prefer the conclusion based on the most specific knowledge, we remark that there is not universal agreement on this in the probabilist community when statistics are not good. Kyburg [7,8] suggests we make an inference based on the narrowest reference class for which we have adequate statistics. Some Bayesians suggest that data from various subclasses be combined[2][2].

**Lemma 4.5** If $b \models a$, but $a \not\models b$ then $conf(b, a)$.

**Proof:** This generalizes the property of logical links to the rest of the graph. □

### 4.4 Transitive inference

Default proofs consist of more than a single inference; part of the appeal of such reasoners is that they appear to create and argument by making inferences towards a goal. In general, if $a \to b$ and $b \to c$ are links on an inference graph, we cannot conclude $conf(c, a)$. However, if $c$ is conditionally independent of $a$ given $b$, it can be shown that $conf(c, a)$. In fact, conditional independence gives us much more than transitivity. Not only can we reverse the inference, we can perform *transduction*, inferring evidence from other evidence. We can also confirm certain conjunctions.

---

[2]This reference was pointed out to us by Peter Cheeseman.

**Lemma 4.6** *(Probabilistic Resolution)* If there exists $c$ such $conf(a,c)$ and $conf(b,c)$ and $a$ is conditionally independent of $b$ given $c$, then $conf(a,b)$.

**Proof:** By contradiction. Suppose $p(a|b) \leq p(a)$. From the premises and an identity of probability it follows $p(a|b) \leq p(a|\neg b)$. Then,

$$p(a|b) = p(a|cb)p(c|b) + p(a|\neg cb)p(\neg c|b).$$
$$p(a|\neg b) = p(a|c\neg b)p(c|\neg b) + p(a|\neg c\neg b)p(\neg c|\neg b).$$

Simplify using the conditional independence knowledge, then subtract to obtain

$$\begin{aligned} 0 &\leq p(a|\neg b) - p(a|b) \\ &= (p(a|c) - p(a|\neg c))(p(c|\neg b) - p(c|b)). \end{aligned}$$

But then both terms must be positive, contradicting the premise that $conf(b, c)$. □

Unsurprisingly, each such inference results in a dilution of confirmation. This lemma is needed for later results.

**Lemma 4.7** If $conf(a,b)$ and $conf(b,c)$ and $a$ is conditionally independent of $c$ given $b$, then $p(a|c) < p(a|b)$.

**Proof:**

$$\begin{aligned} p(a|c) &= p(a|bc)p(b|c) + p(a|\neg bc)p(\neg b|c) \\ &= p(a|b)p(b|c) + p(a|\neg b)p(\neg b|c) \\ &< p(a|b)p(b|c) + p(a|b)p(\neg b|c) \\ &= p(a|b)(p(b|c) + p(\neg b|c)) \\ &= p(a|b) \square \end{aligned}$$

The next two lemmas yield two ways of confirming conjunctions of events.

**Lemma 4.8** *(Irrelevance)* If $conf(a,c)$ and $a$ is conditionally independent of $b$ given $c$, then $conf(a,bc)$.

**Proof:** $p(a|bc) = p(a|c) > p(a)$, from the definition of conditional independence. □

**Lemma 4.9** *(Relevance)* Suppose $conf(a,c)$ and $conf(b,c)$ and $a$ is conditionally independent of $b$ given $c$. Then $conf(ab,c)$.

**Proof:** $conf(a,b)$ follows from Lemma 4.6, and $p(a|b) < p(a,c)$ from Lemma 4.7. Then

$$p(ab|c) = p(a|c)p(b|c) < p(a|b)p(b). \square$$

## 4.5 Other inferences

The following two lemmas address situations that prove to be useful in Section 5.3. The proofs are straightforward and we omit them.

**Lemma 4.10** If $conf(\neg a, b)$, $conf(a, c)$, and $b \Rightarrow c$, then $conf(a, \neg bc)$.

**Lemma 4.11** If $r$ and $e$ are the direct predecessors of $g$, and

1. $r \models e$,
2. $a \models e$,
3. $r$ is unconditionally independent of $a$,
4. $conf(g, e)$,
5. $conf(\neg g, r)$,

then $conf(g, a)$.

## 5 Examples

### 5.1 Birds fly

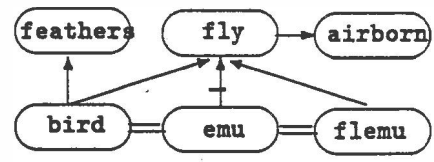

This inference graph aims to capture a lot of information. If something is a bird, we believe it is more likely that it will fly, and if it flies, it is more likely to be airborn. If something is an emu, we are less likely to

279

believe it flies, but if it is a flemu (flying emu) we again change our belief. We are inclined to believe that birds have feathers. What inferences can we make from this graph?

**Birds fly, emus don't** We can prove *conf(fly, bird)* and *conf(¬fly, emu)* from the single links containing this information. More importantly, we cannot prove *conf(fly, emu)*, i.e., we do not have the "multiple extension" problem. We can use Lemma 4.4 to conclude *conf(¬fly, bird∧emu)*.

**Emus don't vanish** Exactly the opposite is true: we show *conf(emu, bird)*. Note that we do not accept conclusions, just increase our belief in them.

**Emus are not airborn** In default reasoners based on theorem provers, there is typically a single proof of *airborn* from *emu* using the default "birds fly". Poole [13] solves this by accepting only what is true in every extension. The semantics of extensions are not very well understood; it is trivial to generate probability distributions where propositions true in every extension are less likely than those that are not. Thus the meaning of this conclusion is unclear, We conclude an individual is less likely to be airborn given it is an emu because *conf(¬airborn, ¬fly)* and *conf(emu, ¬fly)* and *airborn* is conditionally independent of *emu* given *fly*.

**Feathered things fly** With Lemma 4.6, we can show *conf(feathers, fly)* using $c = bird$.

## 5.2 Modified Nixon Diamond

Below is the historic example of *not* wanting to draw an inference. If Dick is both a Quaker and a Republican, we do not want to conclude he is a hawk or dove.

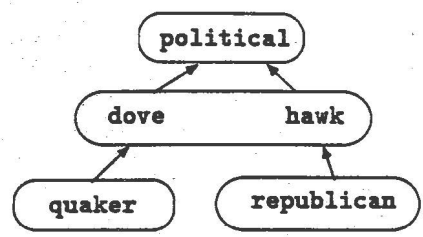

Our system concludes that *quaker* increases belief in *dove* and *republican* increases belief in *hawk*. Since the graph contains no information about the joint distribution, we do not conclude either *hawk* or *dove* if Nixon is both.

However, we want to conclude Quakers and Republicans are *political*. Inheritance systems cannot represent mutual exclusion[5]. Default reasoners simply add all the links with the result that *political* is true given *dove* or *¬dove*[13]. It is possible to prove that an object about which nothing is known is a political non-emu!

We solve the problem in this formalism by making *hawk* and *dove* mutually exclusive but not necessarily exhaustive outcomes of some random event. Since *political* is conditionally independent of *quaker* given *dove*, we can make the desired inference using Lemma 4.6. The price we pay for consistency and transitive inference is not being able to show that *¬hawk* is confirmed by *quaker*. If we allow this, then we can confirm both *political* and its negation given *quaker*.

## 5.3 Royal and African Elephants

This appears in [24,5]. Intuitively, the graph is suppose to show elephants are typically gray, but Royal elephants are not. If Clyde is both and African and a Royal elephant what are we to conclude about grayness?

We use Lemma 4.11 to conclude African elephants are gray. If *royal* is true, then *elephant* is true and the conditional independence assumptions shield *gray* from the effect of *african* and we conclude *conf(¬gray, [royal* *african]).* This would not be true if the links from *royal* and *african* to *elephant* were probabilistic.

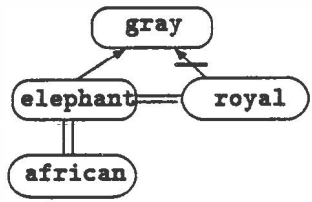

Horty et al and Sandewall disagree on this. We claim there are no "right" answers to this question and we build different graphs to model domains with different properties.

### 5.4 Naive diagnosis

Consider the diagnostic dual to the "birds fly" problem.

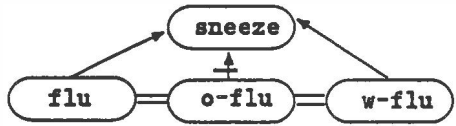

If we observe *sneeze*, a default reasoner produces all three diseases as diagnoses. The inference graph confirms both *flu* and *w-flu*; of these it is easy to prefer the most probable diagnosis confirmed by the observations. If we observe ¬*sneeze* only *o-flu* is confirmed.

## 6 Conclusions

We applied the formalism to many other default inference problems including plan recognition and stereotyping with positive results. In general, we obtain answers that agree with intuition. Where we haven't, the underlying sound probabilistic basis has always provided the tool for understanding the structure of the particular problem.

We have implemented the system in Prolog. A set of input probabilistic and logical arcs are compiled into a graph that is used specifically for testing conditional independence using Pearl's definition of d-separability[11]. The rest of the system consists of a straightforward transcription of the Lemmas in Section 4 into Prolog and the system prints a readable proof of the probabilistic inferences it makes.

## Acknowledgements

The research of the first author was supported by graduate scholarships from NSERC and the Institute for Computer Research at the University of Waterloo. The research of the second author was supported under NSERC grant A6260. Thanks to Romas Aleliunas, André Trudel, Bruce Spencer, Peter VanBeek, Jimmy Lee and Paul van Arragon for thorough comments on an earlier draft. Fahiem Bacchus suggested Lemmas 4.10 and 4.11. Thanks to the anonymous referees for comments.